\documentclass[runningheads]{llncs}
\usepackage{graphicx}
\usepackage{amsmath}
\usepackage{tikz}
\usepackage{ifthen}
\usepackage{makecell}
\usepackage{multirow}
\usepackage{pgfplots}
\usepackage{xcolor}
\usepackage{enumitem}
\usepackage{booktabs}
\usepackage{url}

\usepgfplotslibrary{fillbetween}

\definecolor{darkcyan}{RGB}{66, 215, 244}
\begin{document}
\title{Look Who's Talking: Inferring Speaker Attributes from Personal Longitudinal Dialog}
%
%
\author{Charles Welch \and
Ver\'{o}nica P\'{e}rez-Rosas \and\\
Jonathan K. Kummerfeld \and 
Rada Mihalcea}
%
%
\institute{University of Michigan\\
\email{\{cfwelch,vrncapr,jkummerf,mihalcea\}@umich.edu}}
%
\maketitle              
\begin{abstract}
We examine a large dialog corpus obtained from the conversation history of a single individual with 104 conversation partners. The corpus consists of  half a million instant messages, across several messaging platforms. We focus our analyses on seven speaker attributes, each of which partitions the set of speakers, namely: gender; relative age; family member; romantic partner; classmate; co-worker; and native to the same country. In addition to the content of the messages, we examine conversational aspects such as the time messages are sent, messaging frequency, psycholinguistic word categories, linguistic mirroring, and graph-based features reflecting how people in the corpus mention each other. We present two sets of experiments predicting each attribute using (1) short context windows; and (2)  a larger set of messages. We find that using all features leads to gains of 9-14\% over using message text only.

\end{abstract}

\section{Introduction} \label{sec:intro}

People spend a significant amount of time using social media services such as instant messaging to communicate and keep in touch with others. Over time, conversation history can grow quickly, thus becoming an abundant source of personal data that provides the opportunity to study an individual's communication patterns and social preferences. Analyzing conversations from a single individual rather than conversations from multiple individuals can enable identification of social behaviors that are specific to that individual. Moreover, longitudinal analyses can help us better understand an individual's social interactions and how they develop over time. 

In this work we look at a collection of personal conversations of one of this paper authors' over a five-year span, consisting of nearly half a million messages shared with 104 conversation partners. To address data privacy issues, during the experiments and analyses presented in this paper, the actual message content is only accessible to its owner. We focus our analyses on seven speaker attributes: a ternary attribute for relative age (younger, older, or same age); and six binary attributes reflecting whether somebody is the same gender; a family member; a romantic partner; a classmate; a co-worker; and a native of the same country. We explore the classification of speaker attributes, i.e, the group(s) the speaker belongs to, using a variety of linguistic features, message and time frequency features, stylistic and psycholinguistic features, and graph-based features. In addition, we examine the performance increase gained by using six of the attributes as features to try to classify the seventh.

We analyze linguistic variation in messages exchanged between the author and the other speakers. We also conduct analyses that look at speaker interaction behaviors, considering aspects such as time, messaging frequency, turn-taking, and linguistic mirroring. Next, we apply graph-based methods to model how people interact with each other by representing people as nodes and speaker mentioning each other as directed edges. 
We then apply clustering methods to identify groups that naturally occur in the graph. Finally, we conduct several classification experiments to quantify the impact of features derived from these analyses on our ability to determine who a speaker is. 

Identifying speaker attributes has important applications within the areas of personalization and recommendation \cite{rao2010classifying,garera2009modeling}. While a large number of conversations that occur online are short, such as interactions on Twitter, there are also many social media platforms where personal dialog may span thousands of utterances. For this reason, we conduct evaluations at the level of small context windows, as well as at the speaker level using a large set of messages from each speaker. To the best of our knowledge, this is the first study on speaker attribute prediction using personal longitudinal dialog data that focuses on one person's dialog interactions with many other speakers. 

\section{Related Work}
Our work is related to three main directions of research: authorship attribution, discourse analysis, and speaker attribute classification from social media.

On authorship attribution, there have been several studies focusing on inferring author's characteristics from their writing, including their gender, age, educational, cultural background, and native language~\cite{Hirst07,Koppel09}. This work has considered linguistic features to capture lexical, syntactic, structural, and style differences between individuals~\cite{Koppel09}. A recent study in this area analyzed language use in social media to identify aspects such as gender, age, and personality by looking at group differences on language usage in words, phrases, and topics discussed by Facebook users~\cite{Schwartz13}.

Discourse analysis approaches have been used to examine language to reveal social behavior patterns. Holmer \cite{Holmer08} applied discourse structure analysis to chat communication to identify and visualize message content and interaction structures. He focused on visualizing aspects such as conversation complexity, overlapping turns, distance between messages, turn changes, patterns in message production and references. In addition, he also proposed graph-based methods for showing coherence and thread patterns during the messaging interaction. Tuulos \cite{Tuulos04} inferred social structures in chat-room conversations, using heuristics based on participants' references, message response time and dialog sequences and represented social structure using graph-based methods. Similarly, Jing \cite{jing2007extracting} looked at extracting networks of biographical facts from speech transcripts that characterize the relationships between people and organizations.

Work in classifying user attributes has used both message content and other meta-features. Rao \cite{rao2010classifying} looked at classifying gender, age (older or younger than 30), political leaning, and region of origin (north or south India) as binary variables using a few hundred or a few thousand tweets from each user. They used the number of followers and following users as network information to look at frequency of tweets, replies, and retweets as communication-based features but found no differences between classes. Hutto \cite{Hutto13} analyzed sentiment, topic focus, and network structure in tweeting behavior to understand aspects such as social behavior, message content and following behavior. Other work has derived useful information from Twitter profiles, such as Bergsma \cite{bergsma2013broadly} who focused on gender classification using features derived from usernames, and Argamon \cite{argamon2003gender} who found differences in part of speech and style when examining gender in the British National Corpus.

\section{Conversation Dataset}

We use a corpus of text messages from one author's personal conversations on Google Hangouts, Facebook Messenger, and SMS text messages. The message set contains nearly half a million messages from conversations held between the author and 104 individuals. 
Aggregate statistics describing the corpus are shown in Table \ref{tab:corpus_numbers}.

\begin{table}[!ht]
    \centering
    \small
    \caption{Distribution of messages and tokens (words, punctuation, emoticons) in the conversations between the author and other individuals.}
    \begin{tabular}{rrrr} \toprule
	&	Author	&	Others	&	All 	\\
    \midrule
Total Messages	&	237,300	&	216,766	&	454,066	\\
Unique Messages	&	165,536	&	168,041	&	326,243	\\
Total Tokens	&	1,370,916	&	1,602,607	&	2,973,523	\\
Unique Tokens	&	38,937	&	48,005	&	68,985	\\
Average Tokens / Message	&	5.78	&	7.39    &	6.55		\\ \bottomrule
    \end{tabular}
    
    \label{tab:corpus_numbers}
\end{table}

\begin{table*}[h]
    \centering
    \small
    \caption{Distribution of speakers and messages in the corpus by speaker attributes (\% of corpus). The values for \textit{Age} represent `younger', `older', and `same age', while the values for the other attributes represent `yes' and `no'.}
    \begin{tabular}{lccccccc}
    \toprule
             & & Rom. & Rel. & Child. & & &  \\
             & Family & Rel. & Age & Co. & Gender & School & Work \\ \midrule
         & Y/N & Y/N & Y/O/S & Y/N & Y/N & Y/N & Y/N \\ \midrule
        \%Speakers & 6/94 & 9/91 & 26/30/44 & 78/20 & 51/49 & 62/38 & 33/67 \\
        \%Messages & 8/92 & 22/78 & 24/24/52 & 88/11 & 53/47 & 75/25 & 54/46 \\ \bottomrule
    \end{tabular}
    
    \label{tab:group_numbers}
\end{table*}

We use seven attributes that describe the relationship between the author and their conversation partner. Table \ref{tab:group_numbers} shows the distribution of people and messages for each attribute in the dataset. They were annotated by the author and interpreted as follows:

\begin{description}[noitemsep] 
    \item[Family:] This person is related to the author.
    \item[Romantic Relationship (Rom. Rel.):] This person's relationship with the author was at some point not platonic.
    \item[Relative Age (Rel. Age):] This person the same age ($\pm 1.5$ years), is older, or is younger than the author.
    \item[Childhood Country (Child. Co.):] This person grew up in the same country as the author.
    \item[Gender:] This person has the same gender as the author.
    \item[School:] This person and the author met attending school.
    \item[Work:] This person and the author know each other because they worked together.
\end{description}

\section{Message Content} 

We start by exploring linguistic differences in the messages exchanged between the author and each of the groups defined by the seven attributes described above. 
We obtain the most dominant semantic word classes~\cite{pulman2009linguistic} in messages exchanged with people sharing each attribute using the LIWC~\cite{tausczik2010psychological} lexicon, which contains psycholinguistic categories of words. The top ten dominant classes for each attribute-value pair are shown in Table \ref{tab:ling_ethnography}.

\begin{table}[!ht]
    \centering
    \small
    \caption{Dominant LIWC word classes for each attribute/value pair. The top ten classes are listed for each attribute in decreasing order.}
    \begin{tabular}{p{1.88cm}p{9.4cm}}
    \toprule
    Attribute & Top Classes \\ \midrule
    Family & \textbf{Yes:} Family, Money, Home, Swear, Death, Leisure, Filler, Anger, Female, Health \\
    & \textbf{No:} Anxious, Insight, Feel, Risk, Sad, Positive Emotion, Non-fluencies, Causality, Affect, Work \\
    Romantic Relationship & \textbf{Yes:} Anxious, Death, Sad, Feel, Body, Filler, You, Family, Perception, Health \\
     & \textbf{No:} Swear, Female, Money, Friend, Anger, She-He, Work, Leisure, Informal, Male \\
    Relative Age & \textbf{Younger:} Netspeak, Ingest, Swear, Friend, Biological, Home, Anger, Informal, Body, Leisure \\
    & \textbf{Same:} Female, Swear, Anger, She-He, Anxious, Negative Emotion, Friend, Sad, Negate, Money \\
    & \textbf{Older:} See, We, Work, Number, Article, Home, Perception, Space, Motion, Relativity \\
    Childhood Country & \textbf{Same:} Death, Family, Anger, Swear, Feel, Female, Negative Emotion, Body, Anxious, Health \\
    & \textbf{Other:} We, Work, You, Male, Focus Future, Social, Affiliation, Friend, Assent, Time \\
    Gender & \textbf{Yes:} Money, Female, Swear, Work, Friend, Netspeak, She-He, Article, Power \\
    & \textbf{No:} Sad, Anxious, Family, Health, Death, Body, Biological, Negative Emotion, Ingest, Home \\
    School & \textbf{Yes:} Work, Non-fluencies, Insight, Risk, Anxious, Quantify, Focus Past, Causality, Tentative, Compare \\
    & \textbf{No:} Family, Money, Health, Home, Netspeak, Death, Swear, Leisure, Biological, Anger \\
    Work & \textbf{Yes:} Work, Article, Number, We, Non-fluencies, Quantify, Compare, Insight, Achievement, Assent \\
    & \textbf{No:} Family, Health, Money, Death, Anger, Swear, Anxious, Home, Biological, Sad \\ \bottomrule
    \end{tabular}
    
    \label{tab:ling_ethnography}
\end{table}

Not surprisingly, the `Family=Yes' group talks more about family and home than the `Family=No' group. Interestingly, people who are not family members seem to use more emotion related words. Word categories related to feelings are also very dominant for the `Romantic Relationship=Yes', `Relative Age=Same', `Childhood Country=Same' and `Gender=No' groups; however they seem to focus on negative emotions such as anxiety and sadness.
In fact, those two are in the top three classes for conversations with romantic partners `Romantic Relationship=Yes', which also includes death words (words related to death are often used in hyperbole, e.g. ``I didn't eat lunch and I'm dying"). This suggests that more serious conversations occur between the author and this group as compared to the `Romantic Relationship=No' group.

Several of the attributes clearly separate the set of speakers into those who speak about work and those who do not. People who talk the most about work are those who grew up in other countries (`Childhood Country=Other'), people from work (`Work=Yes'), people older than the author (`Relative Age=Older'), people with the same gender (`Gender=Yes') and people from school (`School=Yes'). However, there are some differences between these groups which can be seen mostly in the family, health, time, and gender specific words they use.

People from school use more words referring to the past, while people from other countries focus more on the future. Interestingly, people not from work (`Work=No') and the people not from school (`School=No') are very similar, and both use a lot of family, health, and money words. The similarity of these two attributes is also interesting in that people from work (`Work=Yes') and/or school (`School=Yes') use more quantifying words (e.g. sampling, percent, average) and disfluencies (e.g. umm, hmm, sigh). We also see that those who grew up in other countries use more male words, while speakers that are the same age, from the same country, or of the same gender use more female words.

\section{Groups Over Time}
\begin{figure}[!ht]
\begin{tikzpicture}
    \centering
    \begin{axis}[
        height=5cm, width=12.5cm,
        yshift=5.3cm, xshift=0cm,
        title style={font=\small},
        label style={font=\small},
        tick label style={font=\small},
        ymin=0.11, 
        legend style={font=\tiny},
        legend pos=south west,
        ymajorgrids=true,
        xtick={0,1,2,3,4,5,6},
        xticklabels={M,Tu,W,Th,F,Sa,Su},
        legend cell align={left},
    ]
        \addplot[
            color=black,
            mark=oplus,
            line width=0.3mm,
        ]
        coordinates {(0,0.1592976632)(1,0.1524963353)(2,0.1477213935)(3,0.1431124429)(4,0.1373825127)(5,0.1222234198)(6,0.1377662326)};
        \addlegendentry{All}
        \addplot[
            color=green,
            mark=otimes,
            line width=0.3mm,
        ]
        coordinates {(0,0.1610884273)(1,0.1376628021)(2,0.1311954222)(3,0.1569457277)(4,0.1397788573)(5,0.146633684)(6,0.1266950794)};
        \addlegendentry{Family=Yes}
        \addplot[
            color=darkcyan,
            mark=triangle,
            line width=0.3mm,
        ]
        coordinates {(0,0.1547477363)(1,0.1443601291)(2,0.1369288482)(3,0.1381189957)(4,0.1347818222)(5,0.1389901837)(6,0.1520722848)};
        \addlegendentry{Work=No}
        \addplot[
            color=red,
            mark=square,
            line width=0.3mm,
        ]
        coordinates {(0,0.1585619605)(1,0.1367909988)(2,0.1410681258)(3,0.1434956843)(4,0.1544967633)(5,0.1371763255)(6,0.1284101418)};
        \addlegendentry{Child Country=No}
    \end{axis}
    \begin{axis}[
        height=5cm, width=12.5cm,
        yshift=1.0cm, xshift=0cm,
        title style={font=\small},
        label style={font=\small},
        tick label style={font=\small},
        legend style={font=\tiny},
        legend pos=south east,
        ymajorgrids=true,
        xtick={3,6,9,12,15,18,21,24},
        legend cell align={left},
    ]
        \addplot[
            color=black,
            mark=oplus,
            line width=0.3mm,
        ]
        coordinates {(1,0.070500439)(2,0.0603808688)(3,0.0355094277)(4,0.0186190894)(5,0.0101425645)(6,0.0050085706)(7,0.0034156946)(8,0.0041515114)(9,0.0063882269)(10,0.0104373093)(11,0.0186253606)(12,0.0343199967)(13,0.0430076508)(14,0.0457774154)(15,0.0532003846)(16,0.0585643213)(17,0.0572891843)(18,0.0605104728)(19,0.0556001505)(20,0.0571073205)(21,0.0593816631)(22,0.0695723065)(23,0.0802479201)(24,0.0822421506)};
        \addlegendentry{All}
        \addplot[
            color=green,
            mark=otimes,
            line width=0.3mm,
        ]
        coordinates {(1,0.0479985564)(2,0.0320591862)(3,0.0147965475)(4,0.0060750053)(5,0.0017142342)(6,0.0017443084)(7,0.0019849027)(8,0.0056840396)(9,0.0065561938)(10,0.0090824336)(11,0.0166310788)(12,0.0329012661)(13,0.0501939791)(14,0.0569005443)(15,0.0665243151)(16,0.0758774172)(17,0.0744940002)(18,0.0693512977)(19,0.0679979549)(20,0.0644792638)(21,0.0611710926)(22,0.0823133139)(23,0.0820426453)(24,0.0714264233)};
        \addlegendentry{Family=Yes}
        \addplot[
            color=red,
            mark=square,
            line width=0.3mm,
        ]
        coordinates {(1,0.0747181717)(2,0.0658881921)(3,0.0353391559)(4,0.0152552807)(5,0.0058481782)(6,0.0031164634)(7,0.0021930668)(8,0.0023854411)(9,0.0074833596)(10,0.0069639491)(11,0.0147551075)(12,0.0286060559)(13,0.0393790158)(14,0.0639067369)(15,0.0613289215)(16,0.0731791774)(17,0.0688122812)(18,0.0682928706)(19,0.062156131)(20,0.0617713824)(21,0.0509407102)(22,0.0542495479)(23,0.0673309992)(24,0.0660998038)};
        \addlegendentry{Child Country=No}
        \addplot[
            color=blue,
            mark=diamond,
            line width=0.3mm,
        ]
        coordinates {(1,0.0788097768)(2,0.0754622742)(3,0.0504250797)(4,0.0295749203)(5,0.0172901169)(6,0.0072901169)(7,0.0031243358)(8,0.0021041445)(9,0.0034750266)(10,0.0102019129)(11,0.0216578108)(12,0.0389798087)(13,0.0321785335)(14,0.0326886291)(15,0.0460042508)(16,0.0482465462)(17,0.0530393199)(18,0.0563443146)(19,0.049978746)(20,0.0521360255)(21,0.0579383634)(22,0.0661424017)(23,0.0789054198)(24,0.0880021254)};
        \addlegendentry{Romantic=Yes}
    \end{axis}
    \node[align=center,rotate=90,font=\small] at (-1.0cm, 4.5cm) {Percentage of Messages};
\end{tikzpicture}
\caption{Distribution of messages over time. The top shows the distribution over the day of the week and the bottom shows hour of the day. The groups shown are those that vary the most from the aggregate trend over all speakers.}
\label{fig:time_fig}
\end{figure}
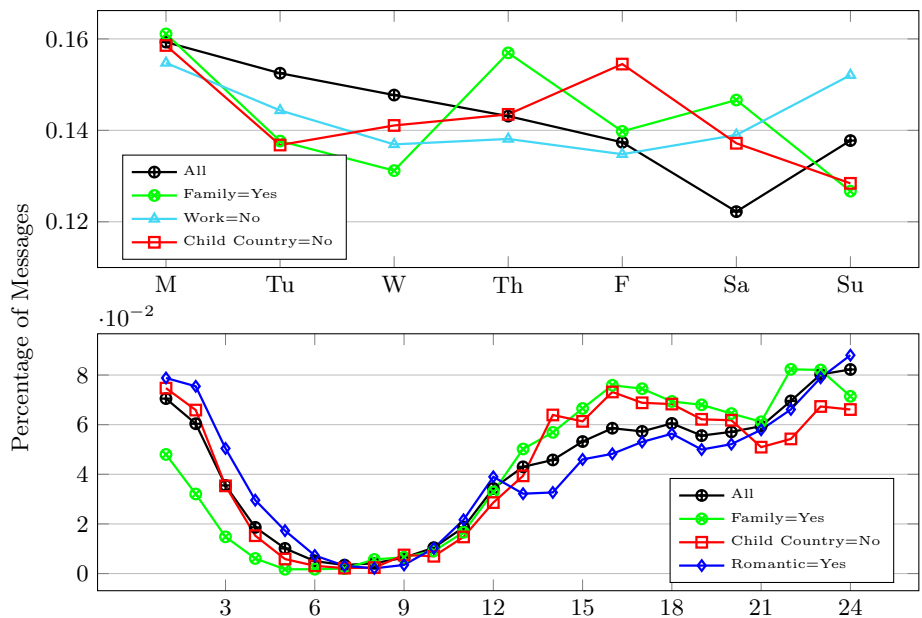

To understand the role that time has in the author's interactions with different groups we look at patterns in message volume over different intervals. Most notably, we find interaction differences given the day of the week, and the hour of the day. In Figure \ref{fig:time_fig} we plot the attribute/value pairs that differ the most from the trend over all people, marked `All'. The difference was calculated as the sum of differences on each of the seven days of the week and each of the 24 hours of the day.

We see that the overall trend for the day-of-week plot (top) is that there are more conversations during the first days of the week. The number of conversations drops until Sunday where it jumps back up and peaks on Monday. Throughout the week, most of the conversations occur between family members and people that grew up in other countries (co-workers mainly). In contrast, there are many more conversations with people outside of work on the weekend.
  

The hour-of-day plot (bottom) indicates that most of the interactions happen between 9AM and 6PM. Though this is a trend aggregated over all days in the corpus it shows that the author is least likely to be talking to people in the 7-8AM range. The author tends to speak more to people later in the day, with a peak at midnight. People who grew up in other countries converse more with the author during the day. The dominant `Work' category for `Childhood Country=Other' in Table \ref{tab:ling_ethnography} shows this trend, as this group may converse with the author more about work during work hours. We also find that family members speak to the author more during the day and romantic partners speak to the author more after midnight but before noon.

\section{Conversation Interaction}\label{sec:interaction}

\begin{figure}[]
	\begin{tikzpicture}
	\centering
	\begin{axis}[
	height=5cm, width=12.5cm,
	title style={font=\small},
	label style={font=\small},
	tick label style={font=\small},
	ymin=0.77, ymax=0.91,
	legend style={font=\tiny},
	legend pos=south east,
	ymajorgrids=true,
	yticklabels={78, 80, 82, 84, 86, 88, 90},
	ytick={0.78, 0.8, 0.82, 0.84, 0.86, 0.88, 0.90},
	xtick={0,1000,2000,3000,4000,5000},
	legend cell align={left},
	]
	\addplot[
	color=black,
	mark=oplus,
	line width=0.3mm,
	]
	coordinates {(100,0.787912678819)(500,0.827554594324)(1000,0.847713455998)(2000,0.875904140959)(3000,0.8876799520449999)(5000,0.8979776763840001)};
	\addlegendentry{Relative Age=Same}
	\addplot[
	color=orange,
	mark=square,
	line width=0.3mm,
	]
	coordinates {(100,0.7883884182740001)(500,0.828944587174)(1000,0.848196770869)(2000,0.876135881436)(3000,0.887285170338)(5000,0.8981793584519999)};
	\addlegendentry{School=Yes}
	\addplot[
	color=red,
	mark=triangle,
	line width=0.3mm,
	]
	coordinates {(100,0.794193757281)(500,0.8325045592150001)(1000,0.853949512658)(2000,0.87856380142)(3000,0.886342358765)(5000,0.893038442296)};
	\addlegendentry{Childhood Country=Same}
	\addplot[
	color=green,
	mark=otimes,
	line width=0.3mm,
	]
	coordinates {(100,0.7729982785699999)(500,0.804579108112)(1000,0.834946352648)(2000,0.880034898082)(3000,0.8894574977659999)(5000,0.8930110985749999)};
	\addlegendentry{Family=Yes}
	\addplot[
	color=blue,
	mark=diamond,
	line width=0.3mm,
	]
	coordinates {(100,0.805620474137)(500,0.853666708924)(1000,0.876290773042)(2000,0.8940438203959999)(3000,0.896160202103)(5000,0.8992982349449999)};
	\addlegendentry{Romantic=Yes}
	\end{axis}
	
	\node[align=center,rotate=90,font=\small] at (-1.0cm, 1.6cm) {Linguistic Style Matching};
	\node[align=center,font=\small] at (5.0cm, -0.8cm) {Number of Messages};
	
	\end{tikzpicture}
	\caption{Language mirroring as a function of the number of messages exchanged within groups. Mirroring is shown over the first 5,000 messages averaged over people in each of the listed groups.
		\label{fig:mirror_fig} }
\end{figure}
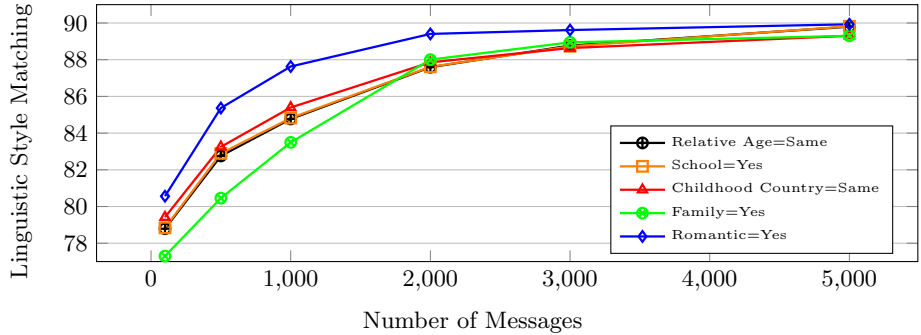

Linguistic mirroring is a behavior in which one person subconsciously imitates the linguistic patterns of their conversation partner. Increased linguistic mirroring can be an indicator of an individual building rapport with others and thus forming better interpersonal relationships. 
We study linguistic mirroring in our dataset to analyze how relationships change over time. We calculate linguistic style matching (LSM) as the similarity of the normalized counts of nine types of function words \cite{Gonzales09}, as the main metric for our analyses. In Figure \ref{fig:mirror_fig} we show style matching over the first 5,000 messages with people in five specific groups. 
We see that although the general trend is to match language style more over time, this trend levels off after 3k messages, potentially because at this point relationships start to consolidate. 

\begin{figure*}[h]
    \begin{tikzpicture}
        \node (img) {\includegraphics[width=1.0\textwidth]{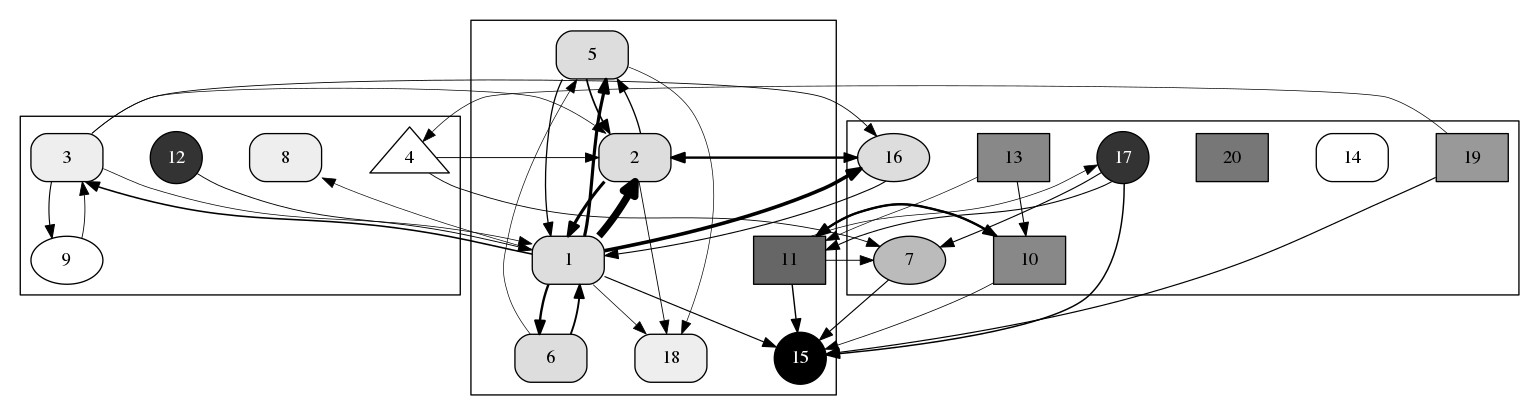}};
        \node [below left,text width=3cm,align=center] at (-2.5cm, -0.3cm) {Cluster 1};
        \node [below left,text width=3cm,align=center] at (1.4cm, 1.5cm) {Cluster 2};
        \node [below left,text width=3cm,align=center] at (6.5cm, -0.3cm) {Cluster 3};
    \end{tikzpicture}
    \caption{Speaker references for the top 20 conversation partners. The graph shows interactions with people from different groups: high school (rectangles), college (triangles), graduate school (rounded rectangles), family members (circles), and other people (ellipses). Shading is proportional to how long ago the author met the person. Edges below a threshold of 25 mentions are removed. Note that the clustering uses all 104 people, but only 20 are shown here.}
    \label{fig:mentions}
\end{figure*}
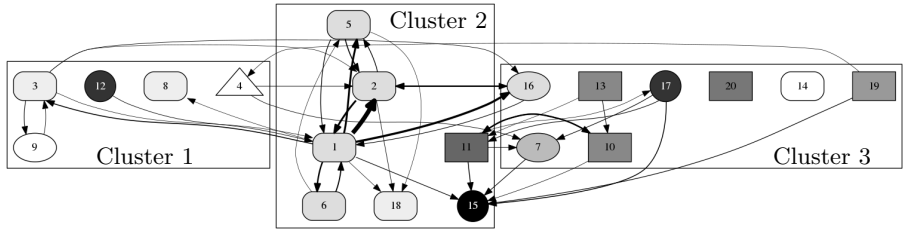

Next, we examine interactions between groups of people by constructing a graph where nodes represent speakers and edges between nodes represent speakers mentioning each other. Speakers who mention each other also tend to know each other. They might mention another person when planning to meet up with others or when talking about an interaction they had with this person in the past. We clustered the graph of people using Louvain clustering \cite{blondel2008fast} to maximize the modularity of the network. This gave four clusters, one of which only contained two people. The remaining clusters roughly evenly split the set of people. The top twenty most frequent conversation partners are shown in Figure \ref{fig:mentions}. 
Interestingly, the clusters resemble groups of speakers that the author spoke most to at three periods of time contained in the corpus i.e, conversations before attending graduate school (Cluster 3), the beginning of graduate school (Cluster 2), and later in graduate school (Cluster 1). We also see that people who spoke to the author more at a particular time were also more likely to know each other.

\begin{table}[]
    \centering
    \small
    \caption{Two examples of five-message context windows ($ctx_1$ and $ctx_2$) taken from the data.} 
    \begin{tabular}{ccp{6.0cm}}
    \toprule
        \multicolumn{1}{l}{Message Number} & \multicolumn{1}{l}{Time} & \multicolumn{1}{l}{Message}\\ \midrule
        $ctx_1msg_0$ & 15:45:06 & Participant: Wanna grab coffee? \\ 
        $ctx_1msg_1$ & 15:45:20 & Author: yeah \\
        $ctx_1msg_2$ & 15:45:25 & Participant: Sweet!!!! \\
        $ctx_1msg_3$ & 15:45:29 & Participant: Meet in the lobby? \\
        $ctx_1msg_4$ & 15:45:52 & Author: okay \\ \midrule
        $ctx_2msg_0$ & 12:21:00 & Participant: Perfect!! \\
        $ctx_2msg_1$ & 15:56:22 & Participant: Wanna go to get Thai? \\
        $ctx_2msg_2$ & 16:01:18 & Participant: I'll take it you're sleeping lol \\
        $ctx_2msg_3$ & 16:19:59 & Author: Yeah \\
        $ctx_2msg_4$ & 16:20:08 & Author: I mean yeah I was sleeping \\ \bottomrule
    \end{tabular}
    
    \label{tab:context}
\end{table}

\section{Model}

Using the messages in a conversation between two speakers, we wish to be able to identify the value of each of the speaker attributes of whom the author is conversing with. In order to do this, we can encode part of the conversation and additional features and output the value of an attribute. In text messaging, it is often not clear what a conversation is about by just examining individual messages. Thus, we decide to conduct our analysis on small sequences of message exchanges between speakers. Throughout the rest of the paper, we will refer to each of these sequences as a \textit{context window}, which consists of five messages exchanged between the author and another speaker\footnote{Context window size is fixed in our experiments but future work could explore prediction accuracy as a function of this variable.}. Two sample context windows are shown in Table \ref{tab:context}.


During our experiments, we use a bidirectional long-short term memory network (BiLSTM) as our baseline model. The input for this model is a dialog context window, in which all utterances are concatenated but one token is used to represent the beginning of an author utterance and another token is used to represent the beginning of any other speaker's utterance. We use the same implementation to incorporate additional features. 

\begin{figure}[t]
\centering
\def\layersep{1.5cm}
\scalebox{0.7}{
\begin{tikzpicture}[shorten >=1pt,->,draw=black!50, node distance=\layersep]
    \tikzstyle{every pin edge}=[<-,shorten <=1pt]
    \tikzstyle{neuron}=[circle,fill=white,minimum size=17pt,draw=black,inner sep=0pt,line width=1pt]
    \tikzstyle{input neuron}=[neuron, fill=green!50];
    \tikzstyle{output neuron}=[neuron, fill=red!40];
    \tikzstyle{hidden neuron}=[neuron, fill=blue!30];
    \tikzstyle{operator}=[neuron];
    \tikzstyle{lstm cell}=[neuron, fill=blue!30, shape=rectangle];
    \tikzstyle{annot} = [text width=4em, text centered]
    
    \pgfdeclarelayer{encoder1}
    \pgfdeclarelayer{encoder2}
    \pgfdeclarelayer{encoder3}
    \pgfdeclarelayer{encoder4}
    \pgfdeclarelayer{decoder1}
    \pgfdeclarelayer{decoder2}
    \pgfdeclarelayer{decoder3}
    \pgfsetlayers{encoder1,encoder2,encoder3,encoder4,decoder1,decoder2,decoder3}

    \begin{pgfonlayer}{encoder1}
        \path[fill=yellow!10,rounded corners, draw=black!50, dashed]
            (-6cm, -5.5cm) rectangle (-0.8cm, 0.3cm);
        \node (eDES) at (-4.5cm, 0cm) {\textit{Context Encoder}};

        \node[input neuron, pin=left:] (I-1) at (-5cm,-1.5) {$w_{1}$};
        \node[input neuron, pin=left:] (I-2) at (-5cm,-2.5) {$w_{2}$};
        \node[input neuron, pin=left:] (I-3) at (-5cm,-3.5) {...};
        \node[input neuron, pin=left:] (I-4) at (-5cm,-4.5) {$w_{n}$};
    
        \node[lstm cell] (F-1) at (\layersep-5cm,-1.75) {$F_1$};
        \node[lstm cell] (F-2) at (\layersep-5cm,-2.75) {$F_2$};
        \node[lstm cell] (F-3) at (\layersep-5cm,-3.75) {...};
        \node[lstm cell] (F-4) at (\layersep-5cm,-4.75) {$F_n$};
        
        \node[lstm cell] (B-1) at (\layersep-4cm,-1.25) {$B_1$};
        \node[lstm cell] (B-2) at (\layersep-4cm,-2.25) {$B_2$};
        \node[lstm cell] (B-3) at (\layersep-4cm,-3.25) {...};
        \node[lstm cell] (B-4) at (\layersep-4cm,-4.25) {$B_n$};
        
        \draw (F-1) -- (F-2) -- (F-3) -- (F-4);
        \draw (B-4) -- (B-3) -- (B-2) -- (B-1);
        \foreach \name / \y in {1,...,4}
            \draw[bend right=10,->] (I-\name) to node [auto] {} (F-\name);
        \foreach \name / \y in {1,...,4}
            \draw[bend left=10,->] (I-\name) to node [auto] {} (B-\name);
        
        \node[output neuron] (ES) at (\layersep-3cm,-4.75) {$\rho_1$};
        \draw (F-4.east) to node [auto] {} (ES.west);
        \draw (B-1.east) .. controls (\layersep-2.75cm,-1.25) and (ES.north) .. (ES.north);
        
        \node[operator] (concat) at (2.5cm,-4.75) {};
        \draw[draw=black,-] (concat.north) to (concat.south);
        \draw[draw=black,-] (concat.west) to (concat.east);
        \draw (ES) -- (concat);
    \end{pgfonlayer}
    
    \begin{pgfonlayer}{encoder2}
        \path[fill=yellow!10,rounded corners, draw=black!50, dashed]
            (0.0cm, -3.6cm) rectangle (4.2cm, 0.3cm);
        \node[output neuron] (FENCx1) at (2.5cm,-3.5) {$\rho_i$};
    \end{pgfonlayer}
    \begin{pgfonlayer}{encoder3}
        \path[fill=yellow!10,rounded corners, draw=black!50, dashed]
            (0.2cm, -3.8cm) rectangle (4.4cm, 0.1cm);
        \node[output neuron] (FENCx2) at (2.5cm,-3.5) {$\rho_i$};
    \end{pgfonlayer}
    \begin{pgfonlayer}{encoder4}
        \path[fill=yellow!10,rounded corners, draw=black!50, dashed]
            (0.4cm, -4.0cm) rectangle (4.6cm, -0.1cm);
        \node (e2DES) at (2.0cm, -0.4cm) {\textit{Feature Encoders}};
        
        \node[input neuron, pin=above:] (II-1) at (1.5cm,-1.5) {$f_{1}$};
        \node[input neuron, pin=above:] (II-2) at (2.5cm,-1.5) {...};
        \node[input neuron, pin=above:] (II-3) at (3.5cm,-1.5) {$f_{t}$};
        
        \node[hidden neuron] (H-1) at (1.0cm,-2.5) {$h_{1}$};
        \node[hidden neuron] (H-2) at (2.0cm,-2.5) {$h_{2}$};
        \node[hidden neuron] (H-3) at (3.0cm,-2.5) {...};
        \node[hidden neuron] (H-4) at (4.0cm,-2.5) {$h_{s}$};
        
        \node[output neuron] (FENC) at (2.5cm,-3.5) {$\rho_i$};
        \draw (FENC) -- (concat);
        
        \foreach \source in {1,...,3}
            \foreach \dest in {1,...,4}
                \path (II-\source) edge (H-\dest);
        \foreach \dest in {1,...,4}
            \path (H-\dest) edge (FENC);
    \end{pgfonlayer}
    
    \begin{pgfonlayer}{decoder1}
        \path[fill=yellow!10,rounded corners, draw=black!50, dashed]
            (5.4cm, -5.1cm) rectangle (8.6cm, 0.3cm);
    \end{pgfonlayer}
    \begin{pgfonlayer}{decoder2}
        \path[fill=yellow!10,rounded corners, draw=black!50, dashed]
            (5.6cm, -5.3cm) rectangle (8.8cm, 0.1cm);
    \end{pgfonlayer}
    \begin{pgfonlayer}{decoder3}
        \path[fill=yellow!10,rounded corners, draw=black!50, dashed]
            (5.8cm, -5.5cm) rectangle (9.0cm, -0.1cm);
        \node[align=right] (dDES) at (8.1cm, -0.7cm) {\textit{\makecell[r]{Attribute\\ Decoders}}};
        
        \node[input neuron] (A-1) at (6.8cm,-1.75) {$c_{1}$};
        \node[input neuron] (A-2) at (6.8cm,-2.75) {$c_{2}$};
        \node[input neuron] (A-3) at (6.8cm,-3.75) {...};
        \node[input neuron] (A-4) at (6.8cm,-4.75) {$c_{s}$};
        
        \draw (concat.east) .. controls (7.5cm,-4.5) and (4.0cm,-1.75) .. (A-1.west);
        \draw (concat.east) .. controls (7.0cm,-4.5) and (5.0cm,-2.75) .. (A-2.west);
        \draw (concat.east) .. controls (6.5cm,-4.5) and (6.0cm,-3.75) .. (A-3.west);
        \draw (concat.east) to node [auto] {} (A-4.west);
        
        \node[output neuron, pin={[pin edge={black,->}]0:}] (AT1) at (7.8cm,-2.25) {$a_1$};
        \node[output neuron, pin={[pin edge={black,->}]0:}] (AT2) at (7.8cm,-3.25) {...};
        \node[output neuron, pin={[pin edge={black,->}]0:}] (AT3) at (7.8cm,-4.25) {$a_k$};
        
        \foreach \source in {1,...,3}
            \foreach \dest in {1,...,4}
                \path (A-\dest) edge (AT\source);
    \end{pgfonlayer}
\end{tikzpicture}
}
\caption{The model architecture encodes a context window as a sequence of tokens $w_1$ to $w_n$ using a BiLSTM which is represented with forward and backward cells. The encoding is then used in combination with our other feature sets for decoding. A separate decoder is used for each speaker attribute.}
\label{fig:model}
\end{figure}
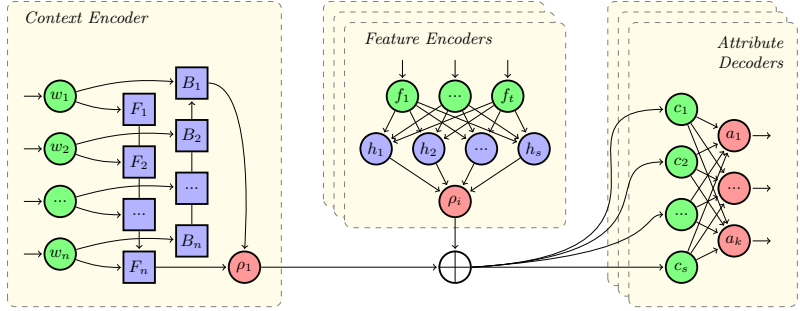

The model architecture is shown in Figure \ref{fig:model}. As shown, the context encoder takes the concatenated window of length $n$ and generates the encoding $\rho_1$.  In the baseline case the feature encoders are not used and the context encoding is passed directly to an attribute decoder. A separate attribute decoder is used for each speaker attribute and has $k$ outputs, where $k$ is two for every case except `relative age', which has three possible values.


When using additional features, we take the BiLSTM output, representing the encoded context window, and append it to a normalized vector representation of each additional feature set, $\rho_i$. A feed-forward layer is then used to encode each feature set separately. The hidden size $s$ for both the feature encoders and attribute decoders were manually tuned in preliminary experiments.

We use a hidden size of 64 for experiments in this paper. In our models that use one or more feature encoders, the concatenated $\rho$ vector is used for decoding. The feature encoder sizes $t$ will vary depending on which feature set is being encoded. The word embedding inputs to the context encoder are 300 dimensional.

\section{Features} \label{sec:features}

\begin{description}
    \item [Word Embeddings:] We obtain word vector representations for each message using the GloVe Common Crawl pre-trained model~\cite{pennington2014glove}. We chose GloVe over other frequently used embeddings because its training data is more similar to our data and we observed a higher token coverage rate than embeddings such as word2vec trained on GoogleNews \cite{mikolov2013distributed}.
    
    \item [LIWC:] To calculate these features, we obtain the normalized counts of 73 LIWC categories. The feature set includes the vectors obtained from messages of individual conversation participants, the cosine similarity between them, and the vector sum of both speakers.

    \item [Time:] These features include the time elapsed during the context window, the number of seconds between each of the messages, and the day, month, year, season (winter, fall, summer, spring), and hour of the day of the last message. 
    
    \item [Messaging frequency:] This set of features includes the number of messages exchanged between conversation participants in the past day, week, month, and from all time. The vector also includes a list of binary values representing the turn change sequence in the context window.
    
    \item [Style Matching:] Looks at the similarity of the ratios of function word usage between the two speakers. This set of features includes the LSM score for the last hundred messages exchanged by the conversation participants, as well as the change in style matching over the context window by subtracting the final and initial LSM scores.
    
    \item [Graph-based:] Uses the training set of messages to generate a graph where nodes represent people and weighted, directed edges represent how often that person mentions another person when speaking to the author. This graph is used to generate features by finding the shortest path between users where edge weights are smaller when they have more mentions. We then use the adjacency matrix to find the shortest paths between nodes and use each row as a feature set, representing a speaker $i$ conversing with this person. Given a graph of mentions, where $M_{i,j}$ represents how often person $i$ mentions person $j$, we compute weights using the following equation:
    \begin{center}$
    \begin{aligned}[t]
      W_{i,j} = 1 - \frac{w_{max} - M_{i,j}}{w_{max} - w_{min}} \\
    \end{aligned}$
    \end{center}
    
    \item [Speaker Attributes:] When we are predicting one of the seven speaker attributes this feature set represents the values of the other six attributes. Note that we cannot use this feature when training joint models.
    
\end{description}

\section{Experiments}
Using the features described in Section \ref{sec:features} we run experiments using leave-one-speaker-out cross validation. We take the 104 speakers in our dataset and hold out all context windows containing dialog with one of the speakers as a test set and use the rest for training and validation with a 90\% and 10\% split. This means that we train and tune parameters on context windows from all 103 other speakers and update the model based on its predictions on each individual context window. During test time we examine the context-level and speaker-level accuracy. Context-level accuracy is calculated by macro-averaging the context window accuracy over all speakers. To calculate accuracy at speaker level, we first obtain the attribute prediction at context-window level for the held-out speaker and assign the attribute value most frequently predicted by the classifier.

We run experiments using a baseline model which only uses word embeddings and compare it to a model that uses all of our features. Additionally, we perform an ablation to examine the effectiveness of each feature set for predicting each speaker attribute by running the model using the word embeddings plus one of the other feature sets at a time. While we vary the number of feature encoders we use (see Figure \ref{fig:model}), each model always uses one attribute decoder. The loss for each model is calculated as the cross-entropy loss for that model's attribute decoder.

Since this evaluation is computationally expensive we run our experiments on a subset of the original corpus. Thus, we obtain a sample of 27,316 context windows, distributed as evenly as possible, from each speaker in the dataset to ensure that all people and attributes are represented. Experiments using this dataset took 3-4 days to run on a cluster with 12 NVIDIA GeForce GTX TITAN X GPUs.

\begin{table*}[ht]
    \centering
    \small
    \caption{Results are shown for the accuracy per person using leave-one-speaker-out cross validation. Individual models learn to classify each attribute in all cases except for the two `Joint' rows, which jointly classify attributes. Feature ablations are shown for each of the single feature types, and compared to the model that uses all features, as well as the baselines obtained using the majority class or message embeddings (Emb) only. Additional improvements are shown when training single attribute classifiers and using the other six attributes as features.}
    \scalebox{1.0}{
    \begin{tabular}{lccccccc}
    \toprule
    \multicolumn{1}{l}{} & \multicolumn{1}{l}{} & \multicolumn{1}{c}{Rom.} & \multicolumn{1}{c}{Rel.} & \multicolumn{1}{c}{Child.} & \multicolumn{1}{c}{} & \multicolumn{1}{l}{} & \multicolumn{1}{l}{} \\
    \multicolumn{1}{l}{} & \multicolumn{1}{c}{Family} & \multicolumn{1}{c}{Rel.} & \multicolumn{1}{c}{Age} & \multicolumn{1}{c}{Co.} & \multicolumn{1}{c}{Gender} & \multicolumn{1}{c}{School} & \multicolumn{1}{c}{Work} \\ \midrule
    \multicolumn{8}{c}{Baselines} \\ \midrule
    Majority Class & 94.2 & 91.3 & 44.2 & 77.9 & 51.0 & 61.5 & 67.3 \\
    Emb & \textbf{94.2} & \textbf{91.3} & 45.2 & 79.8 & \textbf{86.5} & 73.1 & 80.8 \\ \midrule
    \multicolumn{8}{c}{Single Attribute Decoder Ablation} \\ \midrule
    Emb + Time & 94.2 & 91.3 & 44.2 & 79.8 & 85.6 & 76.0 & 85.6 \\
    Emb + LIWC & 94.2 & 91.3 & 46.2 & 80.8 & 82.7 & 73.1 & 84.6 \\
    Emb + Style & 94.2 & 91.3 & 49.0 & 78.8 & 86.5 & 76.0 & 85.6 \\
    Emb + Frequency & 94.2 & 91.3 & 44.2 & 80.8 & 83.7 & 75.0 & 86.5 \\
    Emb + Graph & 93.3 & 91.3 & 43.3 & 77.9 & 80.8 & 76.0 & \textbf{87.5} \\ \midrule
    \multicolumn{8}{c}{Single Attribute Decoder All Features vs Joint Decoder Models} \\ \midrule
    All Features & 92.3 & 91.3 & 45.2 & 81.7 & 76.0 & \textbf{76.9} & 83.7 \\
    Joint + Emb & 94.2 & 91.3 & 48.1 & 78.8 & 85.6 & 71.2 & 83.7 \\
    Joint + All & 92.3 & 91.3 & \textbf{51.9} & \textbf{84.6} & 77.9 & 75.0 & 84.6 \\ \midrule
    \multicolumn{8}{c}{Single Attribute Decoder with Attribute Features} \\ \midrule
    Emb + Attributes & 94.2 & 91.3 & 48.1 & 87.5 & 83.7 & 73.1 & 84.6 \\
    All + Attributes & 93.3 & 91.3 & 50.0 & \textit{\textbf{88.5}} & 78.8 & \textit{\textbf{78.8}} & 85.6 \\ \bottomrule
    \end{tabular}
    }
    
    \label{tab:results1}
\end{table*}

During our experiments we consider single attribute models, which use only one attribute decoder, and joint models, which learn to predict all attributes at the same time using all decoders. In the single attribute setting we train a separate model for each attribute and calculate the cross-entropy loss for the decoder, while in the joint case we take the sum of the losses for all decoders.

\section{Results}

The results obtained for each attribute, when using different combinations of features are shown in Table \ref{tab:results1} and Table \ref{tab:results2}. The first table shows accuracies at the person-level while the latter shows performance macro-averaged over context-windows. Overall, the combination of all features improves the prediction performance for all the attributes over a baseline model that only uses word embeddings, with the exception of the gender attribute. The largest context-window level improvements are obtained for the \textit{Relative age}, \textit{Childhood country}, \textit{Gender} and \textit{Work} attributes. The largest speaker-level improvements are similar with the addition of \textit{School} and without \textit{Gender}.

Although in some cases the accuracy of attribute prediction at speaker-level is not improved by the different set of features, we still observe an improvement on the prediction accuracy at the context window level. For instance, the \textit{Family} and \textit{Romantic} attributes improve by 2.1\% and 6\% respectively. We also see that the \textit{Gender} attribute improves up to 6.8\% by this metric. 

\begin{table*}[ht]
    \centering
    \small
    \caption{Accuracy on context windows macro-averaged over speakers. The individual, joint, single attribute, and baseline models are defined the same way as in Table \ref{tab:results1}.}
    \scalebox{1.0}{
    \begin{tabular}{lccccccc}
    \toprule
    \multicolumn{1}{l}{} & \multicolumn{1}{l}{} & \multicolumn{1}{c}{Rom.} & \multicolumn{1}{c}{Rel.} & \multicolumn{1}{c}{Child.} & \multicolumn{1}{c}{} & \multicolumn{1}{l}{} & \multicolumn{1}{l}{} \\
    \multicolumn{1}{l}{} & \multicolumn{1}{c}{Family} & \multicolumn{1}{c}{Rel.} & \multicolumn{1}{c}{Age} & \multicolumn{1}{c}{Co.} & \multicolumn{1}{c}{Gender} & \multicolumn{1}{c}{School} & \multicolumn{1}{c}{Work} \\ \midrule
    \multicolumn{8}{c}{Baselines} \\ \midrule
    Majority Class & 94.2 & 91.3 & 44.2 & 77.9 & 51.0 & 61.5 & 67.3 \\
    Emb & 92.0 & 86.0 & 39.2 & 75.7 & 63.7 & 64.6 & 69.5 \\ \midrule
    \multicolumn{8}{c}{Single Attribute Decoder Ablation} \\ \midrule
    Emb + Time & 91.7 & 86.8 & 40.5 & 77.4 & 63.4 & 64.4 & 73.1 \\
    Emb + LIWC & 91.9 & 86.4 & 39.6 & 76.7 & 62.6 & 63.8 & 69.4 \\
    Emb + Style & 92.0 & 86.0 & 38.9 & 76.2 & 62.8 & 65.1 & 69.2 \\
    Emb + Frequency & 91.3 & 87.9 & 39.2 & 76.0 & 62.4 & 65.5 & 71.3 \\
    Emb + Graph & 92.1 & 86.2 & 41.7 & 76.9 & 61.4 & 67.2 & 73.3 \\ \midrule
    \multicolumn{8}{c}{Single Attribute Decoder All Features vs Joint Decoder Models} \\ \midrule
    All Features & 92.0 & 88.1 & 42.7 & 78.9 & 61.2 & 67.0 & 76.0 \\
    Joint + Emb & \textbf{93.9} & \textbf{90.9} & 43.4 & 78.0 & \textbf{64.2} & 65.5 & 69.3 \\
    Joint + All & 92.1 & 90.2 & \textbf{47.2} & \textbf{80.8} & 61.8 & \textbf{68.7} & \textbf{78.4} \\ \midrule
    \multicolumn{8}{c}{Single Attribute Decoder with Attribute Features} \\ \midrule
    Emb + Attributes & 92.6 & 86.4 & 41.5 & 84.1 & \textit{\textbf{68.6}} & 72.7 & 78.4 \\
    All + Attributes & 92.0 & 88.2 & 44.3 & \textit{\textbf{85.7}} & 67.1 & \textit{\textbf{74.3}} & \textit{\textbf{83.4}} \\ \bottomrule
    \end{tabular}
    }
    
    \label{tab:results2}
\end{table*}

Using the other six speaker attributes as features to classify the seventh proved to be beneficial in all cases. The graph features also proved useful for all attributes showing gains of up to 6.7\% in speaker-level performance and up to 7\% in context-window level performance. The frequency features gave the biggest performance increase to the \textit{Romantic}, \textit{Childhood country}, and \textit{Work} attributes. Time features improve performance most on \textit{Romantic}, \textit{Gender}, \textit{School}, \textit{Work}.

The overall trend we found in Section \ref{sec:interaction} showed that the most distinct groups when looking at language mirroring were `Family=Yes' and `Romantic=Yes'. However, we found that the language mirroring features that we used, which use a sliding window, were most useful for \textit{Relative age}, \textit{School}, and \textit{Work}. Similarly, LIWC features help for \textit{Relative age} and \textit{Work}, but they also improve prediction performance for \textit{Childhood country} and \textit{Gender}.

At the speaker level, classification is more difficult and we do not see improvement for all attributes when using the additional features or joint decoders. However, at the context-window level we found that joint decoders improved over single attribute decoders in all cases, though using the additional features did not help for \textit{Romantic}, \textit{Family}, and \textit{Gender}. When using single attribute decoding with the other attributes as features we found even higher performance for four of the attributes. Interestingly, \textit{Gender} still does not benefit from using extra features and simply knowing the values of the other speaker attributes gives the best result. The lowest accuracy overall is obtained for relative age, this can be partly explained by the lower baseline as compared to the other attributes, which is influenced by the fact that it has three possible values instead of two.

\section{Conclusion}

In this paper, we addressed the task of classifying the attributes of an individual based on their conversations in a longitudinal dataset. We conducted analyses of several interaction aspects, including message content, speaker groups over time, and interaction during the conversation. We developed a bidirectional LSTM architecture that, in addition to message content, includes a variety of features derived from our analyses, covering the time-stamp of the messages, messaging frequency, psycholinguistic word categories, linguistic mirroring, and graph-based representations of interactions between people. Additionally, to account for scenarios where some attributes are known, we present experiments that evaluate the use of the other six speaker attributes when classifying the seventh.

Our experiments evaluate the accuracy of predictions at the context-window level, which uses only a sequence of five messages for message content, as well as at the speaker level using a larger set of context windows from each speaker. We observed improvements in speaker level accuracy up to 8.7\% and up to 13.9\% accuracy on context windows. We explore the usefulness of each feature with an ablative study and compare two different methods of decoding. For the case of predicting someone's relative age or whether or not they are a co-worker, classmate, or native from the same country, we see improvement at both levels. Our evaluations show improvement over a system that only uses one of these features at a time, as well as over a baseline system that relies exclusively on message content.


To the best of our knowledge, this is the first study on speaker attribute prediction using personal longitudinal dialog data that focuses on one persons' interactions with many users. The code used to extract the conversations from social media, to interactively annotate speakers, and to perform the experiments presented in this paper is publicly available\footnote{\url{https://github.com/cfwelch/longitudinal_dialog}}, so others can conduct analyses on their own data. 

\section*{Acknowledgments}
This material is based in part upon work supported by the Michigan Institute for Data Science, by the National Science Foundation (grant \#1815291), by the John Templeton Foundation (grant \#61156), and by DARPA (grant \#HR001117S0026-AIDA-FP-045). Any opinions, findings, and conclusions or recommendations expressed in this material are those of the author and do not necessarily reflect the views of the Michigan Institute for Data Science, the National Science Foundation, the John Templeton Foundation, or DARPA. 
%
%
%
\bibliographystyle{splncs04}
\bibliography{cicling.bib}

\end{document}